%% file: climatechangeAIws.tex
\documentclass[11pt]{article}
\pdfoutput=1

\usepackage[final]{neurips_2020}

\usepackage[utf8]{inputenc} 
\usepackage[T1]{fontenc}    
\usepackage{hyperref}       
\usepackage{url}            
\usepackage{booktabs}       
\usepackage{amsfonts}       
\usepackage{nicefrac}       
\usepackage{microtype}      
\usepackage{graphicx}
\usepackage{listings}
\usepackage{xcolor}
\usepackage[utf8]{inputenc}
\usepackage{courier}
\usepackage{pythontex}
\input{math_commands.tex}

\definecolor{codegreen}{rgb}{0,0.6,0}
\definecolor{codegray}{rgb}{0.5,0.5,0.5}
\definecolor{codepurple}{rgb}{0.58,0,0.82}
\definecolor{backcolour}{rgb}{0.95,0.95,0.92}

\lstdefinestyle{mystyle}{
    backgroundcolor=\color{backcolour},   
    commentstyle=\color{codegreen},
    keywordstyle=\color{magenta},
    numberstyle=\tiny\color{codegray},
    stringstyle=\color{codepurple},
    basicstyle=\ttfamily\footnotesize,
    breakatwhitespace=false,         
    breaklines=true,                 
    captionpos=b,                    
    keepspaces=true,                 
    numbers=left,                    
    numbersep=5pt,                  
    showspaces=false,                
    showstringspaces=false,
    showtabs=false,                  
    tabsize=2
}

\graphicspath{ {images/} }

\newcommand{\HL}[1]{\textbf{\textit{#1}}} 

\newcommand{\fct}[1]{\mathbf{\texttt{\textbf{#1}}}}

\newcommand{\pymgrid}[1]{\texttt{pymgrid}#1}

\title{pymgrid: An Open-Source Python Microgrid Simulator for Applied Artificial Intelligence Research}

\author{
  Gonzague Henri\thanks{Corresponding author.} \\
  Total EP R\&T \\
  Houston, TX\\
  \texttt{gonzague.henri@total.com} \\
  \And
  Tanguy Levent \\
  Total SE\\
  Palaiseau, France \\
  \And
  Avishai Halev \\
  University of California, Davis \& Total EP R\&T \\
  Davis, CA, USA\\
  \And
  Reda Alami \\
  Total SE\\
  Palaiseau, France \\
  \And
  Philippe Cordier \\
  Total SE\\
  Palaiseau, France \\
}

\begin{document}

\maketitle

\begin{abstract}
    Microgrids – self-contained electrical grids that are capable of disconnecting from the main grid – hold potential in both tackling climate change mitigation via reducing CO$_2$ emissions and adaptation by increasing infrastructure resiliency. Due to their distributed nature, microgrids are often idiosyncratic; as a result, control of these systems is nontrivial. While  microgrid simulators exist, many are limited in scope and in the variety of microgrids they can simulate. We propose \HL{pymgrid}, an open-source Python package to generate and simulate a large number of microgrids, and the first open-source tool that can generate more than 600 different microgrids. \HL{pymgrid} abstracts most of the domain expertise, allowing users to focus on control algorithms. In particular, \HL{pymgrid} is built to be a reinforcement learning (RL) platform, and includes the ability to model microgrids as Markov decision processes. \HL{pymgrid} also introduces two pre-computed list of microgrids, intended to allow for research reproducibility in the microgrid setting.
\end{abstract}

\section{Introduction}
\begin{figure}[h]
  \centering
  \includegraphics[width=0.8\textwidth]{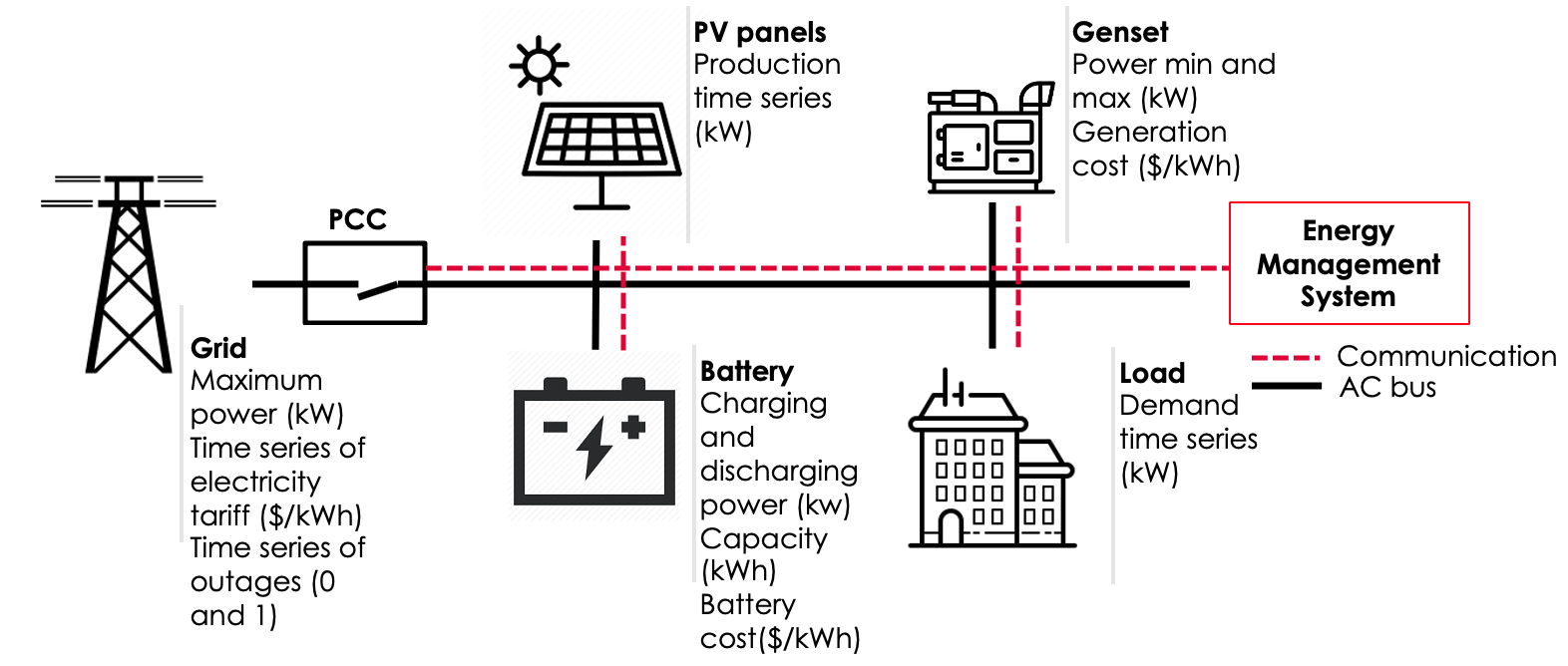}
  \caption{Overview of a microgrid}
  \label{fig:microgrid}
\end{figure}

\vacuum
Microgrids are defined as "a cluster of loads, distributed generation units and energy storage systems operated in coordination to reliably supply electricity, connected to the host power system at the distribution level at a single point of connection, "the point of common coupling" (PCC)"  (Figure \ref{fig:microgrid}) \cite{Olivares2014TrendsControl}. First introduced in 2001, microgrids can also be completely autonomous and disconnected from the grid (off-grid) \cite{Lasseter2001TheConcept}.

Microgrids are one of the few technologies that can contribute to both climate change mitigation - by decreasing greenhouse gas emissions – and adaptation, by increasing infrastructure resiliency \cite{Taft2017ElectricArchitecture} to extreme weather events. Further research is still necessary to fully integrate renewables and reduce costs in order to make widespread microgrid adoptation feasible. Today, one billion people do not have access to electricity; this technology can be used to bring clean energy to communities that are not yet connected to the grid. The importance of bringing clean energy to these communities extends beyond the impact of climate change: indoor pollution, notably the use of dirty fuel for cooking, is a major public health challenge in the developing world \cite{Bruce2000IndoorChallenge}.

Due to their distributed nature, microgrids are heterogeneous and complex systems, potentially equipped with a wide range of generators and employed in a myriad of applications. In addition, as load and renewable generation are stochastic, microgrids require advanced metering and adaptative control to maximize their potential.
Current technical limitations include the amount of solar that can be integrated while the microgrid operates in islanded mode – the controller needs to maintain stability and power quality – as well as the standardization of such controllers, necessary in order to bring down cost and accelerate their deployment \cite{Farrokhabadi2020MicrogridExamples}. Furthermore, as the grid becomes increasingly digital, ever increasing data processing capabilities are required \cite{Abe2011DigitalFuture}. A motivation for this package is to develop tools that can best integrate with the grid of the future.

Microgrid control can be categorized in three main levels. In primary control, voltage and frequency are controlled in a to sub-second time scale. At the secondary level, control focuses on steady state energy control to correct voltage and frequency deviation. Finally,  tertiary control – the focus in \HL{pymgrid} – concerns itself with the long term dispatch of the various generators for optimizing the operational cost of the microgrid. 

In our review of the literature surrounding tertiary control, we observed two main limitations: first, a lack of open-source microgrid models, and second, a lack of a standard dataset to benchmark algorithms across research groups (along with standard performance measurement). Together, these lead to a lack of reproductibility and a difficulty in validating published algorithms.

In this paper, we introduce \HL{pymgrid}, an open-source python package that serves as a microgrid virtual environment.

Through \HL{pymgrid}, we propose two list of pre-compute microgrids, \texttt{pymgrid10} and \texttt{pymgrid25}. Our intention is for them to be used as benchmark scenarios for algorithm development, allowing for more robust research reproductibility.

\vspace{-3mm}\noindent

 \section{Prior Work} \label{prior-work}

Open source Python power systems simulators exist; however, they are often limited in scope \cite{Thurner2018PandapowerSystems, Brown2018PyPSA:Analysis}. Considerations of microgrids in the literature focus on large-scale power systems \cite{Patterson2015HybridApplications, Bonfiglio2017ALogics}. An open-source simulator in the OpenAI gym environment, representing a microgrid for RL, exists, but targets primary control applications \cite{Bode2020TowardsControl, Brockman2016OpenAIGym}. Other models are available on GitHub but either do not simulate tertiary control, are difficult to scale to multiple microgrids or do not allow for straightforward RL integration \cite{DRL-for-microgrid-energy-management, MicrogridRLsimulator,MicroGrids}.
To the best of the authors knowledge,  there does not exist an open source simulator for a large number of microgrids focusing on tertiary control as of September 2020.

As discussed in \cite{RolnickTacklingLearning}, machine learning (ML) algorithms hold promise in power systems and grid related topics. RL and control are listed as relevant for enabling low-carbon electricity. 
Historically, the field has leveraged ML for forecasting, anomaly detection or uncertainty quantification \cite{Bhattarai2019BigDirections}.

However, with recent advances in RL and algorithms beating the world champion of Go, an increasing number of researchers are interested in applying RL to grid-related control applications \cite{Silver2016MasteringSearch}. In \cite{Glavic2017ReinforcementPerspectives,Zhang2018ReviewGrids, Vazquez-Canteli2019ReinforcementTechniques, Mahmoud2017AdaptiveSurvey}, the authors propose reviews of reinforcement learning applications to power systems and the grid. In 2020, the Learning to Run Power Networks Challenge will take place, challening the community to build a Reinforcement Learning agent to manage the real-time operations of a power grid \cite{Kelly2020ReinforcementOperation}. 

\vacuum
\section{Pymgrid} \label{pymgrid}
\vacuum

\HL{pymgrid} consists of three main components: a data folder containing load and PV production time series that are used to 'seed' microgrids, a microgrid generator class named \HL{MicrogridGenerator}, and a microgrid simulator class called \HL{Microgrid}. 

\vacuum
\subsection{Data Collection}
\vacuum
In order to easily generate microgrids, \HL{pymgrid} ships with load and PV production datasets. The load data comes from DOE OpenEI \footnote{\url{https://openei.org}}, and is based on the TMY3 weather data; the PV data is also based on TMY3 and is made available by DOE/NREL/ALLIANCE \footnote{\url{https://rredc.nrel.gov/solar/old_data/nsrdb/1991-2005/tmy3/}}. These datasets contain a year long timeseries with a one hour time-step for a total of 8760 points. Included in the datasets are load and PV files from five cities, each within a different climate zone in the US.

\vacuum
\subsection{Microgrid} \label{pymgrid-microgrid}
\vacuum

This class contains a full implementation of one microgrid; it contains the time series data and the specific sizing for one microgrid. 
\HL{Microgrid} implements three families of functions: the control loop, two benchmark algorithms, and utility functions.

A few functions need to be used in order to interact with a \HL{Microgrid} object. The function $\fct{run()}$ is used to move forward one time step, it takes as argument a control dictionary and returns the updated state of the microgrid. The control dictionary centralizes all the power commands that need to be passed to the microgrid in order to operate each generator at each time-step. Once the microgrid reaches the final time-step of the data, its $done$ argument will pass to $True$. The $\fct{reset()}$ function can be used to reset the microgrid at its initial state, emptying the tracking data structure and resetting the time-step. Once a control action is applied to the microgrid, it will go through checking functions to make sure the commands respect the microgrid constraints.

For reinforcement learning benchmarks and more generally for machine learning, another extra function is useful, $\fct{train\_test\_split()}$ allows the user to split the dataset in two, a training and a testing set. The user can also use $\fct{reset()}$ to go from the training set to the testing set, using the argument $testing = True$ in the reset function. An example is provided in Listing. 

\subsection{MicrogridGenerator class}
\vacuum
\HL{MicrogridGenerator} contains functionality to generate a list of microgrids. For each requested microgrid, the process proceeds as follows. First, the maximum power of the load is generated randomly. A load file is then randomly selected and scaled to the previously generated value. The next step is to automatically and randomly select an architecture for the microgrids. PV and batteries are always present -- this might evolve in the future as we add more components -- and we randomly choose if we will have a diesel generator (genset), a grid, no grid or a weak grid (a grid-connected system with frequent outages). In the case of a weak grid, we also implement a back-up genset; if there is either a grid or a weak grid,  an electricity tariff is randomly selected. The electricity tariffs are generated by \HL{MicrogridGenerator}, and are based on commercial tariffs in California and France.

Once the architecture is selected, the microgrids need to be sized. First, a PV penetration is calculated (as defined by NREL as load maximum power / PV maximum power, in \cite{Hoke2012MaximumFeeders}) and this value is used to randomly scale the selected PV profile. Generated grid sizes are guaranteed to be larger than the maximum load, and the genset provides enough power to fulfill the peak load. Finally, batteries are capable of delivering power for the equivalent of three to five hours of mean load.

Once the different components are selected and sized, \HL{MicrogridGenerator} creates a \HL{Microgrid}. This process is repeated to generate a number of user-requested microgrids.

Overall, with five load files, five PV files, two tariffs, three types of grid, and the binary genset choice, the model can generate more than 600 different microgrid -- before even considering the number of possible different PV penetration levels.

\vacuum
\section{Benchmarks and Discussion} \label{b-and-d}

In addition to the aforementioned capabilities, we proposing two standard microgrid lists: \texttt{pymgrid10} and \texttt{pymgrid25}, containing 10 and 25 microgrids, respectively. \texttt{pymgrid10} has been designed as a first dataset for users new to microgrids. It contains 10 microgrid with the same architecture (PV + battery + genset) and is mostly aimed to gain some intuition for what is happening in the simulation. In \texttt{pymgrid25}, all the possible architectures can be found over 25 microgrids. There are four microgrids with only a genset, three with a genset and a grid, nine with only a grid, and nine with a genset and a weak grid.
As we propose these collections of microgrids as a standardize test set, we also implement a suite of control algorithms as a baseline comparison.  Specifically, we implement four algorithms: rule-based control, model predictive control (MPC), Q-learning, and decision tree (DT) augmented Q-learning. \cite{Levent2019EnergyApproach}. Table \ref{numerical-results-pymgrid25} present the results obtained by running this suite on \texttt{pymgrid25}. By examining the results, we see that MPC with a perfect forecast can be viewed as nearly optimal while the RBC can be viewed as a lower bound as it gives the performance achievable by a simple algorithm. The bold values indicate the best performing non-MPC algorithm.

\begin{table}[h]
\small
  \caption{Numerical results on $\pymgrid{25}$}
  \label{numerical-results-pymgrid25}
  \centering
  \begin{tabular}{lllllll}
    \toprule
    \cmidrule(r){1-2}
    Architecture & Metric (k\$)    & MPC     & Rule-based & Q-learning & Q-learning + DT  \\
    \midrule
    \hline
     All & Mean cost  & 11,643  & 19,265     & 389,234  & \textbf{13,385}    \\
    & Total cost & 291,086  & 481,636   & 9,730,870     & \textbf{334,624}     \\
    \midrule
    Genset only & Mean cost  & 19,722   & 57,398  & 337,385  & \textbf{24,777}    \\
    & Total cost & 78,890  & 229,593   & 1,349,543    & \textbf{99,109}     \\
    \midrule
    Grid only & Mean cost  & 8,150   & \textbf{8,372}     & 383,105  & 8,524    \\
    & Total cost & 73,352  & \textbf{75,350}   & 3,447,945     & 76,718    \\
    \midrule
    Grid + Genset & Mean cost  & 19,107  & \textbf{22,327}     & 480,107  & 22,376    \\
    & Total cost & 57,322  & \textbf{66,982}  & 1,440,322     & 67,130     \\
    \midrule
    Weak grid & Mean cost  & 9,058   & 12,190    & 388,118  & \textbf{10,185}  \\
    & Total cost & 81,522  & 109,711   & 3,493,059     & \textbf{91,666}   \\
    
    \bottomrule
  \end{tabular}
\end{table}

As we can see in  Table \ref{numerical-results-pymgrid25}, there are wide variations in the performance of the algorithms.  Q-learning performs poorly; however,  DT-augmented Q-learning outperforms the RBC -- the mean cost of RBC is approximately 44\% greater than DT-augmented Q-learning -- with most of the difference occuring in edge cases, where RBC performs poorly as it lacks algorithmic complexity. While the DT-augmented Q-learning can perform well, it is still fatally flawed by the necessity of using a discrete action space; this requirement reduces the scope of actions the Q-learner is able to learn, and it is exceedingly difficult to ensure that all possible actions are considered in any given discrete action space. In six of the microgrids, RBC outperforms the DT-augmented Q-learning; this suggests that the DT Q-learning may not be able to consistently exceed an acceptable performance lower bound. This issue generally arises in microgrids with only a grid or with a grid and a genset. 

While the difference between the DT-augmented Q-learning and the MPC often appears to be marginal, it is useful to keep in mind that these 25 microgrids have loads on the order of megawatts; as a result, a 13\% difference can account for \$40 million in additional costs. To fight climate change and integrate more renewables, the technology would need to scale beyond the thousands of system deployed. A few percentage points gained could tremendously reduce the operating costs, thus increasing the importance to improve controller performance. RL being a promising solution to achieve this goal.

\vacuum
\section{Conclusion and Future Work} \label{ccl}
\vacuum
\HL{pymgrid} is a python package that allows researchers to generate and simulate a large number of microgrids, as well as an environment for applied RL research. We establish standard microgrid scenarios for algorithm comparison and reproducibility, and provide performance baselines using both classical control algorithms and reinforcement learning.
In order to improve RL-based microgrid controllers, it is critical to have a universal and adaptable baseline simulator -- a role which \HL{pymgrid} fills. A promising new avenue, is to leverage data generated from multiple microgrids to either increase performance or adaptability.

Immediate plans include the addition of a wider suite of benchmark algorithms, including extensive state of the art reinforcement learning approaches. We also plan to allow for additional microgrid components, more complex use cases, and finer time resolution. In addition, we hope to incorporate the ability to pull real-time data. Finally, functionality surrounding carbon dioxide usage and data is valuable in allowing for users to control for carbon efficiency, along with allowing the user to consider roles that carbon tariffs may play in the future of energy generation.

\medskip

\small
\newpage
\bibliography{references}
\bibliographystyle{unsrtnat}
\end{document}

%% file: math_commands.tex
\usepackage{amsmath,amsfonts,bm}

\usepackage{calrsfs}
\DeclareMathAlphabet{\pazocal}{OMS}{zplm}{m}{n}

\newcommand{\vacuum}{\vspace{-3mm}\noindent}